\documentclass{article}

\usepackage{microtype}
\usepackage{graphicx}
\usepackage{subcaption}
\usepackage{booktabs}
\usepackage{hyperref}

\usepackage[preprint]{icml2026}

\usepackage{multirow}
\usepackage{array}
\usepackage{amsmath}
\usepackage{amsfonts}
\usepackage{amssymb}
\usepackage{mathtools}
\usepackage{dsfont}
\usepackage{url}
\hypersetup{hidelinks,pageanchor=false,pdfsubject={Preprint}}


\icmltitlerunning{FlagGAM: Rule-Basis Generalized Additive Models}

\newcommand{\yes}{\ensuremath{\checkmark}}
\newcommand{\no}{\ensuremath{\times}}
\newcommand{\partialyes}{\ensuremath{\triangle}}
\newcommand{\Ddrop}{D_{\rho}}
\newcommand{\std}[1]{{\scriptsize $\pm$#1}}
\newsavebox{\FGtablebox}
\newcommand{\fitwidth}[2]{%
  \sbox{\FGtablebox}{#2}%
  \ifdim\wd\FGtablebox>#1\relax
    \resizebox{#1}{!}{\usebox{\FGtablebox}}%
  \else
    \usebox{\FGtablebox}%
  \fi
}

\begin{document}

\twocolumn[
  \icmltitle{FlagGAM: Rule-Basis Generalized Additive Models for Explainable Tabular Prediction}

  \begin{icmlauthorlist}
    \icmlauthor{Zijie Zhao}{eecs}
    \icmlauthor{Roy E. Welsch}{sloan}
  \end{icmlauthorlist}

  \icmlaffiliation{eecs}{EECS Department, Massachusetts Institute of Technology, Cambridge, MA, USA}
  \icmlaffiliation{sloan}{Sloan School of Management, Massachusetts Institute of Technology, Cambridge, MA, USA}

  \icmlcorrespondingauthor{Roy E. Welsch}{rwelsch@mit.edu}
  \icmlkeywords{interpretable machine learning, generalized additive models, rule-based learning, robust tabular prediction}

  \vskip 0.3in
]

\printAffiliationsAndNotice{}

\begin{abstract}
Tabular applications often require inspectable prediction rules and stable behavior when records are incomplete. We propose FlagGAM, a rule-basis framework that separates feature-level rule construction from prediction. A Flag Core Module converts numerical and categorical variables into sparse, human-readable univariate bases: threshold flags, category-level flags, tail-deviation bases, and categorical step functions. A default additive head combines these bases as a restricted GAM-style predictor, while the retained sparse rule-basis matrix supports mixed-type classification and regression, feature-specific weighting, and optional flexible heads. On clean benchmarks, additive FlagGAM stays close to modern additive and rule-based baselines on classification and improves over global linear modeling on regression, while remaining less flexible than tree-based predictors. Its clearest advantage appears under deployment-time perturbations: across three classification datasets, FlagGAM has the smallest mean AUROC degradation under missingness and numerical noise. Flexible heads improve absolute accuracy and approach strong tree-based baselines, but should be interpreted as nonlinear predictors over learned rule bases. These results support FlagGAM as a constrained additive rule-basis model for applications that need readable rules and stable behavior with incomplete inputs.
\end{abstract}
\section{Introduction}
\label{sec:introduction}

Tabular prediction is common in healthcare, finance, credit screening, and other high-stakes decision domains. Although deep learning has transformed images, text, and speech, its advantage is less consistent on typical tabular datasets, where tree-based methods often remain very strong baselines \citep{grinsztajn2022tree}. In deployed tabular problems, predictive accuracy alone is also insufficient. Data are often heterogeneous, noisy, incomplete, and collected under operational constraints. Practitioners therefore need models that make competitive predictions while producing explanations that are transparent, auditable, and stable under imperfect inputs.

One response to this challenge is explainable artificial intelligence (XAI), where a black-box model is trained first and explained afterward. Representative post-hoc methods include local surrogate explanations such as LIME \citep{ribeiro2016should} and Shapley-value-based explanations such as SHAP \citep{lundberg2017unified}. These tools are useful for summarizing how complex models behave, but they do not change the underlying prediction rule. For high-stakes decisions, this distinction matters: a post-hoc explanation can support inspection, but it is not the same as a model whose prediction rule is transparent by design. This concern has motivated renewed interest in inherently interpretable, or glass-box, models \citep{rudin2019stop,nori2019interpretml}.

Generalized additive models (GAMs) are a central family of glass-box models because they decompose the prediction into a sum of feature-level contributions \citep{hastie1987generalized}. Recent machine learning work has brought this additive structure back into focus. Explainable Boosting Machines (EBMs) use boosted trees to learn accurate additive shape functions \citep{nori2019interpretml}; Neural Additive Models (NAMs) represent each shape function with a neural network while retaining additivity \citep{agarwal2021neural}; NODE-GAM develops differentiable neural GAMs for interpretable deep learning \citep{chang2022nodegam}; and GAMI-Net uses structured additive neural components and selected interactions to balance accuracy and interpretability \citep{yang2021gami}. These models show that additive learning remains a promising direction for explainable tabular prediction.

However, many modern additive models learn flexible smooth, tree-based, or neural shape functions. Such functions can be visualized and inspected, but they do not necessarily produce concise rules of the kind practitioners often communicate, such as ``risk increases when this laboratory value exceeds a threshold'' or ``this category is associated with a higher-risk subgroup.'' In clinical decision-making, credit screening, and structured operational settings, rule-style explanations are useful because they are easy to audit, compare with domain knowledge, and communicate to non-technical stakeholders.

FlagGAM is also related to basis-expansion and rule-ensemble methods. Multivariate Adaptive Regression Splines (MARS) constructs adaptive spline basis functions, including hinge functions and product terms, for flexible regression \citep{friedman1991multivariate}. RuleFit builds sparse linear models over rules extracted from tree ensembles and was designed to combine rule interpretability with predictive accuracy \citep{friedman2008predictive}. Generalized Linear Rule Models (GLRM) use rule-based features inside generalized linear models and can search over conjunction rules for classification and regression \citep{wei2019glrm}. These methods are close baselines. They can exploit higher-order conjunction rules; FlagGAM instead keeps the default representation univariate and feature-wise additive.

The Univariate Flagging Algorithm (UFA) provides an important step toward rule-based interpretability \citep{sheth2019univariate}. UFA automatically searches for feature-level cutoffs whose tail regions are associated with unusually high or low outcome rates. It converts selected cutoffs into binary flags and summarizes each sample by a low-dimensional count of class-associated flags. This produces simple threshold statements and has shown robustness to missing and noisy data. However, the original UFA formulation is mainly designed for continuous-feature binary classification. It does not directly address categorical predictors, multiclass classification, regression, unequal predictive strengths across selected rules, or the connection between rule discovery and GAM-style additive prediction.

In this paper, we propose \textbf{FlagGAM}, a rule-basis framework for generalized additive modeling on heterogeneous tabular data. The central object in FlagGAM is a full sparse rule-basis matrix $Z(x)$, rather than a compact count of triggered flags. A Flag Core Module converts raw variables into human-readable univariate basis functions, including cutoff flags for numerical classification features, category-level flags for categorical classification features, tail-deviation bases for numerical regression features, and categorical step functions. A default Additive Modeling Head then combines these bases into a restricted GAM-style predictor. When predictive performance is the priority, the same rule-basis representation can be passed to a flexible head, with the caveat that the end-to-end model is then no longer a fully additive GAM.

We make four main contributions. First, we introduce FlagGAM as a rule-basis GAM framework that turns screened threshold and category rules into a full sparse basis matrix $Z(X)$, preserving rule-level interpretability without collapsing evidence into class-count summaries. Second, we extend rule construction to numerical and categorical features, classification and regression tasks, compact feature-specific weighting, and optional flexible prediction heads. Third, we provide training-only, within-feature FDR-screened cutoff construction and rule-level missing-value handling so that all rules are learned from training data only and missing inputs have a clear interpretation. Fourth, we evaluate FlagGAM on clinical, credit-risk, census-income, banking, housing, and scientific measurement benchmarks, with direct comparisons against EBM, RuleFit, GLRM, and tree-based baselines. The results show a consistent trade-off: the default additive model remains competitive with modern interpretable baselines on clean classification tasks, improves over global linear modeling in regression, and shows smaller degradation under missing and noisy feature perturbations.

\begin{table*}[t]
\centering
\small
\setlength{\tabcolsep}{4pt}
\renewcommand\arraystretch{1.12}
\caption{Positioning FlagGAM among interpretable additive and rule-based tabular models.}
\label{tab:positioning}
\fitwidth{\textwidth}{%
\begin{tabular}{llccccc}
\toprule
Method & Rule / basis construction
& \shortstack{Explicit rule\\terms}
& \shortstack{Univariate\\by design}
& \shortstack{Feature-wise\\additive default}
& \shortstack{Mixed-type\\inputs}
& \shortstack{Higher-order\\interactions} \\
\midrule
UFA & Statistical tail cutoffs & \yes & \yes & \no & \no & \no \\
EBM & Boosted one-dimensional shape functions & \partialyes & \yes & \yes & \yes & \partialyes \\
RuleFit & Tree-extracted rules and linear terms & \yes & \no & \no & \yes$^{*}$ & \yes \\
GLRM & Optimized conjunction rules & \yes & \partialyes & \partialyes & \yes$^{*}$ & \yes \\
\textbf{FlagGAM} & Screened cutoffs and category-level bases & \yes & \yes & \yes & \yes & \no \\
\bottomrule
\end{tabular}}

\vspace{0.35em}
{\footnotesize\emph{Note.} \yes{} denotes native/default support; \no{} denotes lack of support in the default formulation; \partialyes{} denotes partial, optional, or variant-specific support. $^{*}$ indicates categorical support through encoding or binarization. UFA is binary-classification only in its original form; the other listed variants support classification and regression. EBM can include interactions, but our EBM baseline uses \texttt{interactions=0}; FlagGAM flexible heads can capture interactions but are not the default additive model.}
\end{table*}

\section{Related Work}
\label{sec:related_work}

\paragraph{Additive and glass-box tabular models.}
GAMs provide a classical additive decomposition of predictions into feature-level effects \citep{hastie1987generalized}. Modern variants such as EBM, NAM, NODE-GAM, and GAMI-Net improve the predictive flexibility of the additive template through boosted, neural, or structured components \citep{nori2019interpretml,agarwal2021neural,chang2022nodegam,yang2021gami}. FlagGAM shares the feature-wise additive goal, but differs by using sparse rule-defined bases rather than dense shape functions. This makes the default model less flexible than models with learned smooth functions or interactions, but gives each selected basis a compact threshold or category-level interpretation.

\paragraph{Rule-based and sparse interpretable models.}
RuleFit constructs rules from tree ensembles and fits sparse linear models over the resulting rules and linear terms \citep{friedman2008predictive}. GLRM learns generalized linear models over optimized rule features and can include higher-order conjunctions \citep{wei2019glrm}. Sparse scoring systems such as SLIM and RiskSLIM learn compact integer-valued risk scores for settings where manual scoring and operational constraints are central \citep{ustun2019risk}. Rules-First classifiers and RRL represent additional lines of work on learning or selecting interpretable rules \citep{cohen2019rulesfirst,wang2021rrl}. Fast sparse generalized linear and additive models provide another route to interpretable sparse prediction \citep{liu2022fastsparse}. FlagGAM differs from these lines by focusing on supervised univariate rule bases and a default feature-wise additive head. This restriction gives up higher-order conjunctions by default, but supports native categorical rule bases, mixed classification/regression tasks, and a simple no-evidence semantics for missing inputs.

\paragraph{UFA and threshold-rule screening.}
UFA searches for outcome-enriched univariate tail regions and summarizes triggered flags into compact class scores \citep{sheth2019univariate}. FlagGAM keeps this emphasis on transparent univariate rules but generalizes the representation: rather than collapsing triggered rules into two counts, it retains the full sparse matrix $Z(X)$, supports categorical features and regression bases, learns task-specific additive heads, and optionally uses flexible heads when predictive performance is the priority.

\section{Methods}
\label{sec:methods}

\subsection{From UFA to FlagGAM}
\label{sec:from_ufa_to_flaggam}

\begin{figure*}[t]
\centering
\includegraphics[width=0.95\textwidth]{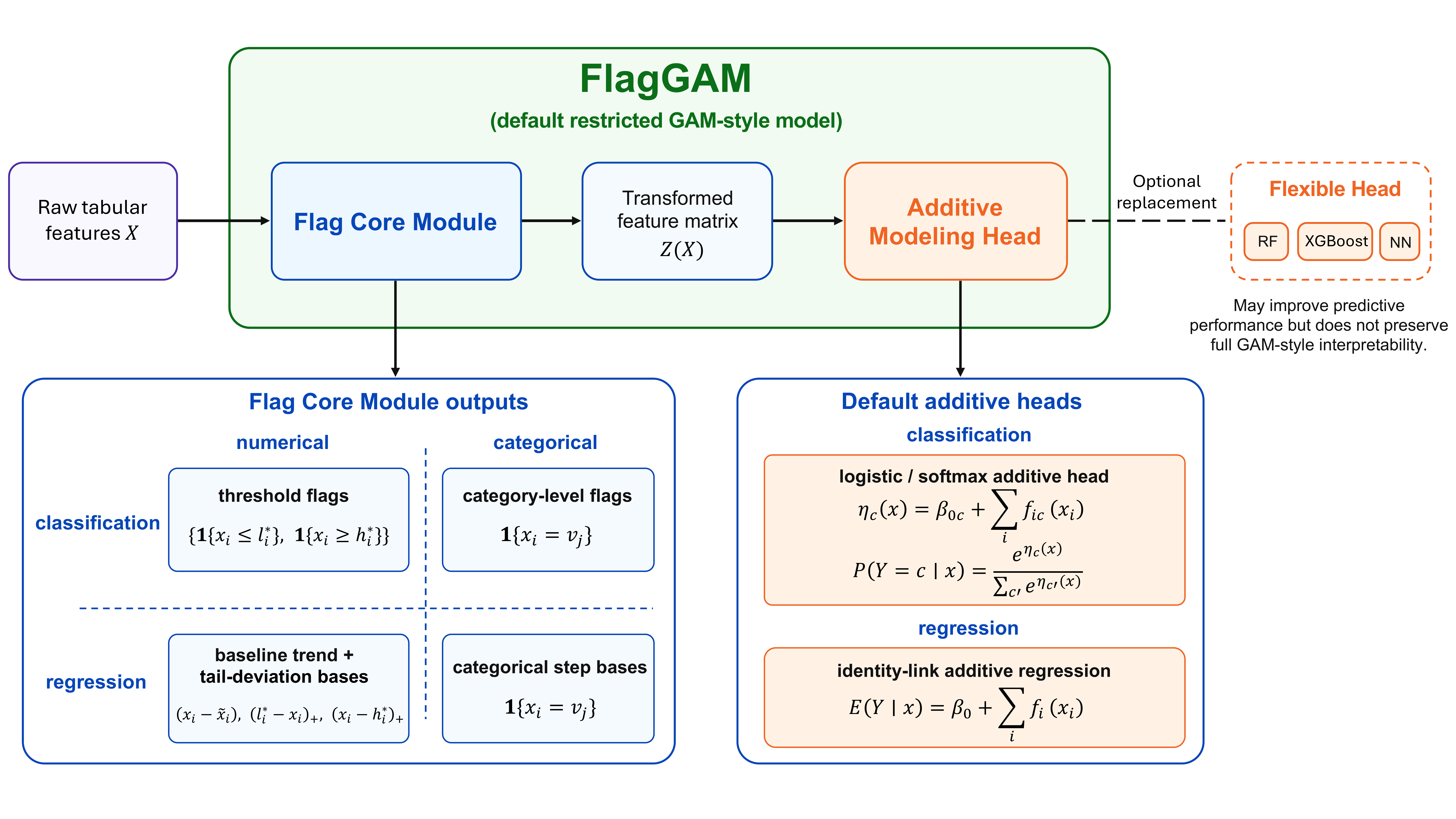}
\caption{FlagGAM pipeline. The Flag Core Module constructs interpretable univariate rule bases from raw tabular features, followed by an additive head or an optional flexible head.}
\label{fig:framework}
\end{figure*}

For a numerical feature, UFA searches for low-tail or high-tail cutoffs whose outcome rate differs from a baseline region. FlagGAM preserves this univariate rule-search logic but uses selected rules as basis functions in a downstream model. Let $x=(x_1,\ldots,x_p)$ be a tabular observation. For each feature $i$, the Flag Core Module constructs a set of basis functions $\mathcal{R}_i=\{z_{ir}(x_i)\}_{r=1}^{m_i}$, where each basis depends only on the original feature $x_i$. The transformed representation is
\begin{equation}
\begin{aligned}
Z(x)=\big[&z_{11}(x_1),\ldots,z_{1m_1}(x_1),\\
&\ldots,z_{p1}(x_p),\ldots,z_{pm_p}(x_p)\big].
\end{aligned}
\end{equation}
The default prediction head is additive over these basis functions. For a generic link function $g$,
\begin{equation}
  g(\mu(x))=\beta_0+\sum_{i=1}^{p} f_i(x_i),
  \qquad
  f_i(x_i)=\sum_{r\in\mathcal{R}_i}\theta_{ir}z_{ir}(x_i).
\label{eq:flaggam_general}
\end{equation}
Because every $z_{ir}$ depends on a single original feature, the default FlagGAM head is a restricted GAM-style model. Figure~\ref{fig:framework} summarizes the FlagGAM pipeline, from feature-level rule-basis construction to additive or optional flexible prediction heads.

\subsection{Flag Core Module}
\label{sec:flag_core}

For numerical classification features, candidate low-tail and high-tail cutoffs are taken from training-set quantiles. A low-tail flag has the form $\mathds{1}\{x_i\le c\}$ and a high-tail flag has the form $\mathds{1}\{x_i\ge c\}$. For binary classification, candidate tails are compared with a baseline region using a two-sided binomial or equivalent proportion test. For multiclass classification, a tail-versus-baseline contingency table is tested using a chi-square statistic. Candidate rules must satisfy a minimum support requirement, and multiple tests are adjusted within each feature using the Benjamini-Hochberg procedure \citep{benjamini1995controlling}. At most one low-tail and one high-tail cutoff are retained per numerical feature; among significant candidates on the same side, FlagGAM chooses the cutoff with the largest effect size.

For categorical classification features, FlagGAM constructs category-level flags $\mathds{1}\{x_i=v\}$ for levels with sufficient support and statistically screened outcome enrichment. For regression, numerical features use a centered baseline trend together with selected tail-deviation bases, such as $(x_i-c)_+$ or $(c-x_i)_+$ after feature centering. Categorical regression features use screened category step functions. In all cases, screening and basis construction use training data only.

\subsection{Additive and flexible heads}
\label{sec:additive_head}

For classification, the default head uses additive class scores. In the multiclass case,
\begin{equation}
\begin{aligned}
\eta_c(x)&=\beta_{0c}+\sum_i f_{ic}(x_i),\\
P(Y=c\mid X=x)&=\frac{\exp(\eta_c(x))}{\sum_{c'}\exp(\eta_{c'}(x))}.
\end{aligned}
\label{eq:classification_softmax}
\end{equation}
Binary logistic regression is recovered as the two-class special case. For regression, the default head uses the identity link, $E(Y\mid X=x)=\beta_0+\sum_i f_i(x_i)$. In both cases, the default model is additive over original features through the selected rule bases.

The same $Z(X)$ representation can also be used by a flexible head such as a Random Forest or boosted tree. This variant may improve predictive accuracy, but it no longer preserves full GAM-style additive interpretation. Throughout the paper, \emph{FlagGAM} refers to the default additive head unless explicitly stated as \emph{FlagGAM + RF Head}.

\subsection{Feature weighting, missing values, and interpretation boundaries}
\label{sec:practical_considerations}

The original UFA count representation treats every triggered flag equally. For compact classification scores, FlagGAM optionally uses feature-specific weights based on training-set association strength. For numerical features with a binary label, we use the absolute point-biserial correlation; for multiclass labels, the correlation ratio; and for categorical features, Cramer's $V$. The main default FlagGAM models in the benchmark tables use the full $Z(X)$ matrix rather than the compact score representation.

Missing values have a simple rule-level interpretation. During rule discovery, each feature-specific screening test uses the available non-missing observations for that feature. At prediction time, a missing numerical value does not trigger threshold flags or tail-deviation bases, and a missing categorical value does not trigger observed category-level flags. Thus, the default behavior is ``missing = no rule evidence,'' rather than forcing the sample into an imputed tail or category. If missingness itself is informative and satisfies support and screening criteria, an explicit missing-indicator flag can be included.

The statistical tests used in the Flag Core Module are screening tools rather than confirmatory post-selection inference. The learned cutoffs and category-level rules should therefore be interpreted as data-adaptive predictive rules rather than formal causal or diagnostic thresholds. FlagGAM also uses univariate feature-level rules by design. This improves transparency, but interactions are not modeled explicitly under the default Additive Modeling Head.

\section{Experiments}
\label{sec:experiments}

\subsection{Experimental protocol}
\label{sec:experimental_protocol}

\begin{table*}[t]
\centering
\small
\renewcommand\arraystretch{1.08}
\caption{Representative clinical rule bases selected by FlagGAM on the Cirrhosis Prediction Dataset. Reference ranges are shown only for clinical context.}
\label{tab:cirrhosis_thresholds}
\fitwidth{\textwidth}{%
\begin{tabular}{llllcc}
\toprule
Variable & Medical interpretation & Ref. range & Rule & Support & Stage-IV rate / lift \\
\midrule
Bilirubin [mg/dL] & high bilirubin & 0.1--1.2 & $\geq 5.94$ & 62 & 58.1\% / 1.66$\times$ \\
Albumin [g/dL] & low albumin & 3.5--5.0 & $\leq 2.97$ & 42 & 59.5\% / 1.70$\times$ \\
Platelets [$10^3/\mu$L] & thrombocytopenia & 150--400 & $\leq 137$ & 41 & 68.3\% / 1.95$\times$ \\
Prothrombin [sec] & higher PT & 9.4--12.5 & $\geq 11.5$ & 73 & 65.8\% / 1.88$\times$ \\
\bottomrule
\end{tabular}}
\end{table*}

\begin{table*}[t]
\centering
\small
\setlength{\tabcolsep}{4pt}
\renewcommand\arraystretch{1.10}
\caption{Classification performance on six tabular benchmarks. Values are AUROC ($\uparrow$), reported as mean $\pm$ standard deviation over 1000 repeated stratified splits.}
\label{tab:classification_auroc}
\fitwidth{\textwidth}{%
\begin{tabular}{lcccccc}
\toprule
Method & Pima $\uparrow$ & Breast Cancer $\uparrow$ & Heart Disease $\uparrow$ & German Credit $\uparrow$ & Adult $\uparrow$ & Bank Marketing $\uparrow$ \\
\midrule
Logistic Regression & 0.836\std{0.024} & 0.985\std{0.011} & 0.865\std{0.027} & 0.758\std{0.029} & 0.903\std{0.007} & 0.784\std{0.007} \\
EBM & 0.855\std{0.020} & 0.991\std{0.007} & 0.895\std{0.022} & 0.792\std{0.024} & 0.921\std{0.005} & 0.807\std{0.006} \\
RuleFit & 0.852\std{0.020} & 0.989\std{0.008} & 0.889\std{0.024} & 0.789\std{0.025} & 0.917\std{0.006} & 0.805\std{0.006} \\
GLRM & 0.849\std{0.021} & 0.990\std{0.008} & 0.886\std{0.025} & 0.787\std{0.026} & 0.919\std{0.005} & 0.809\std{0.006} \\
\textbf{FlagGAM} & 0.848\std{0.021} & 0.988\std{0.009} & 0.880\std{0.025} & 0.775\std{0.026} & 0.912\std{0.006} & 0.801\std{0.006} \\
\midrule
FlagGAM + RF Head & 0.862\std{0.019} & 0.993\std{0.006} & 0.902\std{0.021} & 0.807\std{0.023} & 0.924\std{0.005} & 0.817\std{0.005} \\
RF & 0.858\std{0.020} & 0.986\std{0.010} & 0.891\std{0.023} & 0.785\std{0.025} & 0.918\std{0.005} & 0.811\std{0.006} \\
XGBoost & 0.866\std{0.018} & 0.992\std{0.007} & 0.908\std{0.020} & 0.815\std{0.022} & 0.928\std{0.004} & 0.821\std{0.005} \\
\bottomrule
\end{tabular}}
\end{table*}

For classification tasks, we use 1000 repeated stratified 80/20 train-test splits and report AUROC. For regression tasks, we use 1000 repeated random 80/20 train-test splits and report RMSE and $R^2$. Repeated experiments use seeds indexed from 0 to 999. Within each repeat, all methods use the same train-test split; stochastic models use the repeat seed as the model random state when applicable. All preprocessing, rule discovery, cutoff selection, feature weighting, hyperparameter choices, and model fitting are performed using training data only within each split.

Classification datasets include Pima Indian Diabetes \citep{smith1988using}, Wisconsin Breast Cancer \citep{bennett1992robust}, UCI Heart Disease \citep{janosi1989heart}, Statlog German Credit \citep{hofmann1994statlog}, Adult Census Income \citep{adult1996adult}, and Bank Marketing \citep{bankmarketing2014}. For Bank Marketing, we use the full additional version and remove the \texttt{duration} field because it is only known after the call and is not available at prediction time. Regression datasets include Ames Housing \citep{decock2011ames}, California Housing \citep{californiahousing1997}, and White Wine Quality \citep{cortez2009wine}. For Ames Housing, the target is log sale price and RMSE is computed on this transformed target. Details are provided in Appendix~\ref{appendix:implementation_details}.

We compare the default FlagGAM Additive Modeling Head with Logistic Regression or Ridge Regression, EBM, RuleFit, GLRM, Random Forest, XGBoost, and the FlagGAM + RF Head variant. We treat EBM, RuleFit, and GLRM as the closest interpretable baselines; RF and XGBoost are flexible tabular references.

\subsection{Interpretable rule discovery on clinical tabular data}
\label{sec:clinical_rules}

We first examine whether the Flag Core Module recovers clinically interpretable feature-level rules. We use the Cirrhosis Prediction Dataset, which contains laboratory measurements and histologic disease stages from patients with primary biliary cirrhosis \citep{cirrhosis_uci,murtaugh1994primary}. We convert the four-stage histologic outcome into a binary advanced-stage task by defining stage IV as the positive class and stages I--III as the reference group. In the analyzed samples, the overall stage-IV rate is 35.0\%.

Table~\ref{tab:cirrhosis_thresholds} shows representative rules aligned with common liver-disease markers: high bilirubin, low albumin, low platelets, and prolonged prothrombin time. For this qualitative inspection, rules are learned on a representative training split, and support and stage-IV rate are computed on that analyzed split. Reference ranges are shown only as clinical context and are not used as algorithmic cutoffs. The learned cutoffs are generally more extreme than standard reference-range boundaries because FlagGAM searches for enriched high-risk tails rather than clinical reference limits. These rules are not diagnostic thresholds; they show the type of compact clinical pattern selected by the Flag Core Module.

\begin{table*}[t]
\centering
\small
\setlength{\tabcolsep}{4pt}
\renewcommand\arraystretch{1.10}
\caption{Regression performance on three tabular benchmarks. Values are RMSE ($\downarrow$) and $R^2$ ($\uparrow$), reported as mean $\pm$ standard deviation over 1000 repeated splits.}
\label{tab:regression_performance}
\fitwidth{\textwidth}{%
\begin{tabular}{lcc cc cc}
\toprule
& \multicolumn{2}{c}{Ames Housing} & \multicolumn{2}{c}{California Housing} & \multicolumn{2}{c}{Wine Quality} \\
\cmidrule(lr){2-3}\cmidrule(lr){4-5}\cmidrule(lr){6-7}
Method & RMSE $\downarrow$ & $R^2$ $\uparrow$ & RMSE $\downarrow$ & $R^2$ $\uparrow$ & RMSE $\downarrow$ & $R^2$ $\uparrow$ \\
\midrule
Ridge Regression & 0.173\std{0.012} & 0.844\std{0.020} & 0.736\std{0.014} & 0.591\std{0.014} & 0.748\std{0.019} & 0.287\std{0.031} \\
EBM & 0.128\std{0.009} & 0.916\std{0.012} & 0.552\std{0.009} & 0.771\std{0.008} & 0.675\std{0.016} & 0.419\std{0.027} \\
RuleFit & 0.126\std{0.009} & 0.919\std{0.012} & 0.556\std{0.010} & 0.766\std{0.009} & 0.688\std{0.017} & 0.397\std{0.028} \\
GLRM & 0.132\std{0.010} & 0.911\std{0.014} & 0.568\std{0.010} & 0.756\std{0.009} & 0.681\std{0.017} & 0.410\std{0.028} \\
\textbf{FlagGAM} & 0.142\std{0.010} & 0.895\std{0.015} & 0.585\std{0.010} & 0.743\std{0.009} & 0.686\std{0.017} & 0.400\std{0.028} \\
\midrule
FlagGAM + RF Head & 0.131\std{0.009} & 0.912\std{0.013} & 0.515\std{0.008} & 0.799\std{0.007} & 0.671\std{0.015} & 0.426\std{0.026} \\
RF & 0.136\std{0.010} & 0.905\std{0.014} & 0.497\std{0.008} & 0.813\std{0.007} & 0.666\std{0.015} & 0.434\std{0.025} \\
XGBoost & 0.116\std{0.008} & 0.932\std{0.010} & 0.462\std{0.007} & 0.838\std{0.006} & 0.652\std{0.014} & 0.458\std{0.024} \\
\bottomrule
\end{tabular}}
\end{table*}

\begin{table*}[t!]
\centering
\small
\setlength{\tabcolsep}{3.2pt}
\renewcommand\arraystretch{1.10}
\caption{Robustness under test-time missingness and numerical noise. Values are paired AUROC drops from clean to corrupted test inputs; lower is better.}
\label{tab:robustness_auroc}
\fitwidth{\textwidth}{%
\begin{tabular}{lcc cc cc cc}
\toprule
& \multicolumn{2}{c}{Heart Disease} & \multicolumn{2}{c}{Adult} & \multicolumn{2}{c}{Bank Marketing} & \multicolumn{2}{c}{Mean} \\
\cmidrule(lr){2-3}\cmidrule(lr){4-5}\cmidrule(lr){6-7}\cmidrule(lr){8-9}
Method & 25\% $\downarrow$ & 50\% $\downarrow$ & 25\% $\downarrow$ & 50\% $\downarrow$ & 25\% $\downarrow$ & 50\% $\downarrow$ & 25\% $\downarrow$ & 50\% $\downarrow$ \\
\midrule
\multicolumn{9}{l}{\textit{Missing values}} \\
Logistic Regression & 0.041\std{0.021} & 0.102\std{0.033} & 0.031\std{0.005} & 0.078\std{0.008} & 0.044\std{0.008} & 0.100\std{0.011} & 0.039 & 0.093 \\
EBM & 0.028\std{0.016} & 0.058\std{0.024} & 0.019\std{0.004} & 0.042\std{0.006} & 0.026\std{0.005} & 0.056\std{0.008} & 0.024 & 0.052 \\
RuleFit & 0.030\std{0.018} & 0.064\std{0.026} & 0.023\std{0.004} & 0.052\std{0.006} & 0.032\std{0.006} & 0.073\std{0.009} & 0.028 & 0.063 \\
GLRM & 0.029\std{0.017} & 0.060\std{0.025} & 0.021\std{0.004} & 0.049\std{0.006} & 0.030\std{0.006} & 0.064\std{0.009} & 0.027 & 0.058 \\
\textbf{FlagGAM} & 0.022\std{0.013} & 0.045\std{0.018} & 0.013\std{0.003} & 0.029\std{0.005} & 0.019\std{0.004} & 0.043\std{0.007} & 0.018 & 0.039 \\
FlagGAM + RF Head & 0.027\std{0.014} & 0.054\std{0.021} & 0.017\std{0.003} & 0.038\std{0.006} & 0.024\std{0.005} & 0.052\std{0.008} & 0.023 & 0.048 \\
RF & 0.061\std{0.026} & 0.121\std{0.035} & 0.035\std{0.006} & 0.084\std{0.009} & 0.053\std{0.009} & 0.122\std{0.012} & 0.050 & 0.109 \\
XGBoost & 0.028\std{0.016} & 0.055\std{0.023} & 0.021\std{0.004} & 0.046\std{0.006} & 0.029\std{0.006} & 0.066\std{0.009} & 0.026 & 0.056 \\
\midrule
\multicolumn{9}{l}{\textit{Noisy data}} \\
Logistic Regression & 0.019\std{0.011} & 0.053\std{0.018} & 0.010\std{0.003} & 0.026\std{0.005} & 0.017\std{0.004} & 0.043\std{0.007} & 0.015 & 0.041 \\
EBM & 0.017\std{0.010} & 0.037\std{0.016} & 0.009\std{0.003} & 0.021\std{0.004} & 0.013\std{0.003} & 0.030\std{0.006} & 0.013 & 0.029 \\
RuleFit & 0.020\std{0.012} & 0.044\std{0.017} & 0.011\std{0.003} & 0.028\std{0.005} & 0.016\std{0.004} & 0.038\std{0.007} & 0.016 & 0.037 \\
GLRM & 0.019\std{0.011} & 0.041\std{0.016} & 0.010\std{0.003} & 0.026\std{0.005} & 0.015\std{0.004} & 0.035\std{0.007} & 0.015 & 0.034 \\
\textbf{FlagGAM} & 0.013\std{0.009} & 0.030\std{0.014} & 0.007\std{0.002} & 0.016\std{0.004} & 0.010\std{0.003} & 0.024\std{0.005} & 0.010 & 0.023 \\
FlagGAM + RF Head & 0.016\std{0.010} & 0.036\std{0.015} & 0.009\std{0.002} & 0.021\std{0.004} & 0.013\std{0.003} & 0.030\std{0.006} & 0.013 & 0.029 \\
RF & 0.033\std{0.017} & 0.066\std{0.022} & 0.018\std{0.004} & 0.040\std{0.006} & 0.028\std{0.006} & 0.060\std{0.009} & 0.026 & 0.055 \\
XGBoost & 0.017\std{0.010} & 0.043\std{0.017} & 0.011\std{0.003} & 0.025\std{0.005} & 0.017\std{0.004} & 0.040\std{0.007} & 0.015 & 0.036 \\
\bottomrule
\end{tabular}}
\end{table*}

\subsection{Classification performance}
\label{sec:classification_results}

Table~\ref{tab:classification_auroc} compares default FlagGAM with additive, rule-based, and tree-based baselines. The default additive FlagGAM remains competitive with modern additive and rule-based baselines on clean classification benchmarks while preserving explicit univariate rule bases. The FlagGAM + RF Head variant improves absolute AUROC and approaches the strongest tree-based baselines, showing that the learned rule-basis matrix $Z(X)$ can also serve as a performance-oriented representation when full additive interpretability is not required.

\subsection{Regression performance}
\label{sec:regression_results}

Table~\ref{tab:regression_performance} shows that FlagGAM extends naturally beyond classification. On Ames Housing, FlagGAM reduces RMSE relative to ridge regression while retaining a feature-wise additive form. On California Housing and Wine Quality, the default additive FlagGAM outperforms the global linear baseline but remains below more flexible EBM, RuleFit, GLRM, and tree ensembles on clean regression metrics. This is consistent with its design: the default model preserves univariate rule interpretability rather than explicitly modeling higher-order interactions. The RF-head variant again shows that $Z(X)$ is a useful predictive representation even when the final predictor is not additive.

\subsection{Robustness under missing and noisy data}
\label{sec:robustness_results}

We evaluate deployment-time robustness under test-time missingness and noisy numerical inputs on Heart Disease, Adult, and Bank Marketing. Each model is trained once on clean training data within each split and evaluated on clean and corrupted test sets. For each repeat, all methods share the same clean test set and the same corrupted raw test matrices. In the missing setting, a fraction $\rho\in\{25\%,50\%\}$ of test feature entries is randomly masked. In the noisy-data setting, a fraction $\rho$ of numerical test entries is selected uniformly at random and perturbed by mean-zero Gaussian noise with standard deviation $0.5\sigma_j$, where $\sigma_j$ is the training-set standard deviation of feature $j$. Categorical features are not perturbed by Gaussian noise. For baselines that require complete inputs, missing numerical values are imputed by training-set medians and missing categorical values by training-set modes unless model-native missing handling is used.

For each split $s$, method $m$, dataset, perturbation type, and corruption level, we compute the paired AUROC drop
\begin{equation}
  \Ddrop^{(s)}(m)=\mathrm{AUROC}^{(s)}_{\mathrm{clean}}(m)-\mathrm{AUROC}^{(s)}_{\mathrm{corrupted},\rho}(m).
\end{equation}
Table~\ref{tab:robustness_auroc} reports the mean and standard deviation of these paired drops across repeats, with mean columns computed as unweighted macro-averages of dataset-level mean drops. Lower values indicate smaller degradation. The table measures sensitivity to corrupted inputs; it does not claim that the model with the smallest drop always has the highest absolute AUROC under corruption.

Table~\ref{tab:robustness_auroc} shows that FlagGAM has the smallest mean AUROC degradation under both missing values and noisy numerical inputs. At the 50\% corruption level, the default additive FlagGAM reduces the mean AUROC drop from 0.052 for EBM and 0.056 for XGBoost to 0.039 under missingness, and from 0.029 for EBM and 0.036 for XGBoost to 0.023 under numerical noise. This follows from the rule encoding: missing values contribute no rule evidence, and numerical perturbations affect a rule only when they change whether a selected threshold or tail-deviation basis is activated. These comparisons measure degradation rather than absolute corrupted-test performance; Appendix~\ref{appendix:absolute_corrupted} reports the corresponding 50\% corrupted absolute AUROC values.

\section{Discussion and limitations}
\label{sec:discussion}

FlagGAM targets a different point in the tabular-model design space from flexible tree ensembles: it prioritizes explicit univariate rule bases, feature-wise additive structure, and stable behavior when deployment inputs are incomplete. RuleFit and GLRM can exploit higher-order conjunction rules and may obtain higher clean-data metrics on some tasks. FlagGAM instead preserves a univariate additive rule structure with explicit rule bases and a no-evidence semantics for missing inputs. The robustness results show smaller degradation under the tested missingness and numerical-noise perturbations.

Several limitations follow from this design. First, the statistical tests in FlagGAM are used for rule screening rather than confirmatory post-selection inference, so learned cutoffs should be interpreted as data-adaptive predictive rules rather than formal causal or diagnostic thresholds. Second, the default FlagGAM head does not model higher-order interactions unless they are engineered as features or captured by a flexible head. Third, our robustness study is limited to random test-time missingness and numerical noise; other data-quality shifts are outside the scope of this paper.

Taken together, the results support FlagGAM as a constrained additive rule-basis model: it keeps the main prediction rule additive and inspectable, supports mixed-type classification and regression, and shows smaller degradation under the tested missingness and noisy numerical inputs.

\section*{Impact Statement}
FlagGAM may be useful in tabular workflows where users inspect feature-level rules, such as healthcare or credit screening. The selected rules are learned from data and may reflect bias, measurement artifacts, or sampling patterns. They should not be treated as causal or diagnostic thresholds without domain validation.

\bibliographystyle{icml2026}
\bibliography{references}

\clearpage
\onecolumn
\appendix

\section{Implementation details}
\label{appendix:implementation_details}

Table~\ref{tab:implementation_details} gives the implementation protocol. Hyperparameters are selected using a validation split carved from the training portion: classification settings use validation AUROC and regression settings use validation RMSE. The test split is not used for tuning. After selection, each method is refit on the full training split before test evaluation.

\begin{table}[!htbp]
\centering
\small
\renewcommand\arraystretch{1.04}
\caption{Training-only experimental protocol and baseline tuning. All preprocessing, rule discovery, model selection, and fitting use training data only within each split.}
\label{tab:implementation_details}
\begin{tabular}{p{0.22\textwidth}p{0.70\textwidth}}
\toprule
Component & Setting \\
\midrule
Splits and seeds & 1000 repeated 80/20 train-test splits; classification splits are stratified. Split seeds are 0--999, and stochastic model seeds match the split seed when applicable. \\
FlagGAM rule discovery & Quantile grid uses 5\%--45\% and 55\%--95\% training quantiles with step 5\%; minimum support is $\min(200,\max(20,\lceil0.02n_{\mathrm{train}}\rceil))$; BH-FDR is applied within each feature with $\alpha=0.05$. \\
Categorical and missing values & FlagGAM uses native category-level bases and rule-level no-evidence encoding for missing inputs. Complete-input baselines use training-set median/mode imputation unless model-native missing handling is used. \\
Robustness protocol & Models are trained on clean training data and evaluated on clean and corrupted test sets. Within each repeat, all methods share the same corrupted raw test matrices. Numerical noise uses mean-zero Gaussian perturbations with standard deviation $0.5\sigma_j$; categorical features are not noise-perturbed. \\
Datasets & Bank Marketing excludes \texttt{duration}; Ames target is log sale price; California Housing and Wine Quality use standard regression targets. \\
\midrule
Logistic / Ridge and FlagGAM additive head & $L_2$ regularization is selected from $C\in\{0.01,0.1,1,10\}$ for classification and $\lambda\in\{10^{-3},10^{-2},10^{-1},1,10\}$ for regression. Logistic/Ridge use standardized numerical features and one-hot categorical features; FlagGAM uses the learned $Z(X)$ rule-basis matrix. \\
EBM & InterpretML default main-effect settings with \texttt{interactions}=0, keeping EBM additive for comparison with default FlagGAM. \\
RuleFit & Tree-derived rules with sparse linear/logistic head. Tree depth is selected from $\{2,3,4\}$, maximum rule count from $\{100,200,500\}$, and sparse regularization from the training-only validation split. \\
GLRM & Generalized linear rule model with first-degree and higher-order rules. Maximum rule degree is selected from $\{1,2,3\}$, with sparsity/complexity regularization selected using training data only. \\
RF and RF head & 500 trees, bootstrap enabled, unrestricted max depth, and random state tied to the split seed. The RF head uses only the learned $Z(X)$ basis matrix and no raw features. \\
XGBoost & Depth $\in\{3,4,6\}$, learning rate $\in\{0.03,0.1\}$, and number of trees selected by validation performance or early stopping on training-only validation data. \\
\bottomrule
\end{tabular}
\end{table}

\section{Component ablation}
\label{appendix:component_ablation}

Table~\ref{tab:component_ablation} ablates FlagGAM components on the three robustness datasets. All variants use rules learned by the Flag Core Module but differ in representation and prediction head: compact class scores with equal weights, compact class scores with feature weights, the full $Z(X)$ matrix with the default additive head, or the full $Z(X)$ matrix with an RF head. Missing and noisy evaluations follow the same clean-train, corrupted-test protocol as Table~\ref{tab:robustness_auroc}, and summary statistics are computed over repeated macro-averages across the three datasets. Unlike Table~\ref{tab:robustness_auroc}, this table reports absolute AUROC, because its goal is to assess component utility rather than degradation.

\begin{table}[!htbp]
\centering
\small
\setlength{\tabcolsep}{5pt}
\renewcommand\arraystretch{1.05}
\caption{Component ablation of FlagGAM representations and heads. Values are macro-averaged AUROC across Heart Disease, Adult, and Bank Marketing.}
\label{tab:component_ablation}
\begin{tabular}{p{0.40\textwidth}ccc}
\toprule
Variant & Clean $\uparrow$ & Missing $\uparrow$ & Noisy $\uparrow$ \\
\midrule
Compact scores, equal weights   & 0.839\std{0.012} & 0.787\std{0.016} & 0.814\std{0.013} \\
Compact scores, feature weights & 0.851\std{0.011} & 0.805\std{0.014} & 0.827\std{0.012} \\
Full $Z(X)$, additive head      & 0.864\std{0.010} & 0.825\std{0.012} & 0.841\std{0.011} \\
Full $Z(X)$, RF head            & 0.881\std{0.008} & 0.833\std{0.011} & 0.852\std{0.010} \\
\bottomrule
\end{tabular}
\end{table}

The ablation supports the representation design: feature weighting improves compact scores, the full $Z(X)$ basis improves over compact summaries, and the RF head improves AUROC when additivity is not required.

\section{Sensitivity analysis}
\label{appendix:sensitivity}

Table~\ref{tab:sensitivity_bank} provides a compact sensitivity analysis on Bank Marketing. One rule-discovery parameter is varied at a time around the default setting, while all other settings follow Appendix~\ref{appendix:implementation_details}. Models are trained on clean training data and evaluated on clean, 50\% missing, and 50\% noisy test inputs. The goal is to check predictive sensitivity to reasonable rule-discovery choices, not to optimize each setting separately.

\begin{table}[!htbp]
\centering
\small
\setlength{\tabcolsep}{6pt}
\renewcommand\arraystretch{1.06}
\caption{Sensitivity of FlagGAM rule-discovery hyperparameters on Bank Marketing. Values are AUROC ($\uparrow$) under clean, missing, and noisy test inputs.}
\label{tab:sensitivity_bank}
\begin{tabular}{llccc}
\toprule
& & \multicolumn{3}{c}{Bank Marketing} \\
\cmidrule(lr){3-5}
Varied parameter & Value & Clean $\uparrow$ & Missing $\uparrow$ & Noisy $\uparrow$ \\
\midrule
\multirow{3}{*}{FDR level $\alpha$}
& 0.01 & 0.797\std{0.006} & 0.756\std{0.009} & 0.776\std{0.008} \\
& 0.05 (default) & 0.801\std{0.006} & 0.758\std{0.009} & 0.777\std{0.008} \\
& 0.10 & 0.803\std{0.006} & 0.756\std{0.009} & 0.776\std{0.008} \\
\midrule
\multirow{3}{*}{Quantile-grid step}
& 2.5\% & 0.802\std{0.006} & 0.758\std{0.009} & 0.777\std{0.008} \\
& 5\% (default) & 0.801\std{0.006} & 0.758\std{0.009} & 0.777\std{0.008} \\
& 10\% & 0.797\std{0.007} & 0.755\std{0.010} & 0.775\std{0.008} \\
\midrule
\multirow{3}{*}{Minimum support $s_{\min}$}
& 100 samples & 0.803\std{0.006} & 0.756\std{0.009} & 0.776\std{0.008} \\
& 200 samples (default) & 0.801\std{0.006} & 0.758\std{0.009} & 0.777\std{0.008} \\
& 400 samples & 0.798\std{0.007} & 0.758\std{0.009} & 0.777\std{0.008} \\
\bottomrule
\end{tabular}
\end{table}

Table~\ref{tab:sensitivity_bank} shows that predictive performance is stable across reasonable rule-discovery settings. Relaxing the FDR level or lowering the support threshold slightly changes clean AUROC, but does not consistently improve corrupted-test AUROC. The default setting therefore provides a balanced choice across clean and corrupted-test conditions.

\section{Absolute corrupted-test AUROC}
\label{appendix:absolute_corrupted}

Table~\ref{tab:absolute_corrupted_auroc} reports absolute AUROC values under 50\% corrupted test data for the same robustness datasets used in Table~\ref{tab:robustness_auroc}. It complements the degradation view in the main text; clean AUROC is reported in Table~\ref{tab:classification_auroc}. Thus, smaller degradation does not necessarily imply the highest absolute corrupted-test AUROC.

\begin{table}[!htbp]
\centering
\small
\setlength{\tabcolsep}{4pt}
\renewcommand\arraystretch{1.03}
\caption{Absolute corrupted-test performance under 50\% missingness and 50\% numerical noise. Values are AUROC ($\uparrow$), reported as mean $\pm$ standard deviation.}
\label{tab:absolute_corrupted_auroc}
\fitwidth{\textwidth}{%
\begin{tabular}{lcc cc cc}
\toprule
& \multicolumn{2}{c}{Heart Disease} & \multicolumn{2}{c}{Adult} & \multicolumn{2}{c}{Bank Marketing} \\
\cmidrule(lr){2-3}\cmidrule(lr){4-5}\cmidrule(lr){6-7}
Method & 50\% Missing $\uparrow$ & 50\% Noisy $\uparrow$ & 50\% Missing $\uparrow$ & 50\% Noisy $\uparrow$ & 50\% Missing $\uparrow$ & 50\% Noisy $\uparrow$ \\
\midrule
Logistic Regression & 0.763\std{0.045} & 0.812\std{0.038} & 0.825\std{0.010} & 0.877\std{0.008} & 0.684\std{0.014} & 0.741\std{0.011} \\
EBM & 0.837\std{0.034} & 0.858\std{0.030} & 0.879\std{0.007} & 0.900\std{0.006} & 0.751\std{0.009} & 0.777\std{0.008} \\
RuleFit & 0.825\std{0.033} & 0.845\std{0.031} & 0.865\std{0.008} & 0.889\std{0.007} & 0.732\std{0.011} & 0.767\std{0.009} \\
GLRM & 0.826\std{0.033} & 0.845\std{0.031} & 0.870\std{0.008} & 0.893\std{0.007} & 0.745\std{0.010} & 0.774\std{0.008} \\
\textbf{FlagGAM} & 0.835\std{0.031} & 0.850\std{0.030} & 0.883\std{0.008} & 0.896\std{0.007} & 0.758\std{0.009} & 0.777\std{0.008} \\
FlagGAM + RF Head & 0.848\std{0.029} & 0.866\std{0.028} & 0.886\std{0.008} & 0.903\std{0.006} & 0.765\std{0.008} & 0.787\std{0.007} \\
RF & 0.770\std{0.038} & 0.825\std{0.036} & 0.834\std{0.010} & 0.878\std{0.008} & 0.689\std{0.014} & 0.751\std{0.011} \\
XGBoost & 0.853\std{0.031} & 0.865\std{0.029} & 0.882\std{0.007} & 0.903\std{0.006} & 0.755\std{0.009} & 0.781\std{0.008} \\
\bottomrule
\end{tabular}}
\end{table}

\end{document}